\documentclass{svproc}
\usepackage{url}

\usepackage{multicol}
\usepackage{footmisc}
\usepackage{cite}
\usepackage{amsmath,amssymb,amsfonts}
\usepackage{graphicx}
\usepackage{textcomp}
\usepackage{xcolor}
\graphicspath{ {./Images/} }
\usepackage{hyperref}

\usepackage{ulem}

\usepackage{algorithm}
\usepackage{algpseudocode}
\usepackage{booktabs}
\usepackage{tabularx}
\usepackage{bm}
\usepackage[utf8]{inputenc}

\begin{document}
\mainmatter              
%
\title{A Dynamic Architecture for Task Assignment and Scheduling for Collaborative Robotic Cells}
\titlerunning{A dynamic architecture for task assignment}  
%
\author{Andrea Pupa \and Chiara Talignani Landi \and
Mattia Bertolani \and Cristian Secchi}
\authorrunning{Andrea Pupa et al.} 
%
%
\institute{Department of Science and Method of Engineering, University of Modena and Reggio Emilia, Reggio Emilia, Italy,\\
\email{andrea.pupa@unimore.it}\\
\vspace{0.5cm}
\textbf{\ackname} This project has received funding from the European Union’s Horizon 2020 research and innovation programme under grant agreement No. 818087 (ROSSINI).}

\maketitle              

\begin{abstract}
In collaborative robotic cells, a human operator and a robot share the workspace in order to execute a common job, consisting of a set of tasks. A proper allocation and scheduling of the tasks for the human and for the robot is crucial for achieving an efficient human-robot collaboration. In order to deal with the dynamic and unpredictable behavior of the human and for allowing the human and the robot to negotiate about the tasks to be executed, a two layers architecture for solving the task allocation and scheduling problem is proposed. The first layer optimally solves the task allocation problem considering nominal execution times. The second layer, which is reactive, adapts online the sequence of tasks to be executed by the robot considering deviations from the nominal behaviors and requests coming from the human and from robot. The proposed architecture is experimentally validated on a collaborative assembly job.
\end{abstract}

	\section{Introduction}
	
Collaborative robots are getting more and more common and appealing for industrial applications (see e.g. \cite{villani, bauer}). One of the reasons of this success is the possibility of creating a synergy between a human operator and a robot, two agents with complementary skills, for allowing the execution of jobs that neither the robot nor the human alone can execute. In order to achieve such an ambitious goal it is necessary to enable the collaboration as much as possible by creating mutual awareness and communication between human and robot. A lot of work has been done in this direction for the execution of a specific task. In \cite{Chiara-iros, Serena} human motion prediction strategies have been developed for making the robot aware of the behavior of the human and to react accordingly. In \cite{Anca, bea} the concept of legibility has been exploited for making the intentions of the robot clearly understandable by the human who can, therefore, react accordingly. In \cite{peternel} the authors exploit  electromyography (EMG) signals to estimate the human muscle fatigue. Subsequently, this information is used to adapt the robot behavior, improving the human-robot interaction. In \cite{st} a verbal feedback strategy is implemented. The robot informs the human about what action it will take with verbal information (self-narrative feedback) and what action the human should perform (role-allocative feedback). Moreover the verbal feedback is also exploited to inform the human about the result of a task (empathetic feedback). Instead, in \cite{villanidroni} the authors propose the use of a wearable device to recognize the movements of the human's forearm. This recognition is then used as input to control a semi-autonomous quadrotor.

          In the industrial practice, collaborative cells are exploited for executing a job, composed by a set of tasks. Thus, it is important to understand how to allocate and schedule all the tasks to the human and to the robot (see e.g. \cite{tsarouchi}). 
	
The multi-agent task allocation problem has been widely studied in the industrial scenario. In \cite{blum} the authors implement an algorithm for solving the assembly line worker assignment and balancing problem (ALWABP). The idea is to find the tasks to be assigned to each work station, thus worker, in order to minimize the cycle time respecting the precedence constraints. In \cite{sabattini} the authors use the Hungarian Algorithm to optimally assign a mission between the AGVs, taking into account the traffic state.
	
The problem of task allocation has been addressed also for the human-robot collaboration in industrial scenarios. Several works formulate the task allocation problem as an optimization problem where the human characteristics are encoded in the cost functions (see e.g.\cite{human1,human2,human3, lamon2019}). The main problem with these approaches is the computational complexity, which makes them unsuitable for fast rescheduling. In \cite{bogner} an integer linear programming (ILP) problem for a collaborative assembly scenario is implemented. Due to the complexity of the problem, a metaheuristic approach is used to solve the optimization problem.
In \cite{chen2013} a genetic based revolutionary algorithm for real-time task allocation is proposed. The framework takes into account both parallel and sequential task, while it aims and minimizing the total makespan and the payment cost.
	In \cite{johannsmeier} a framework based on two level of abstraction and allocation in HRC scenario is presented. The first layer is responsible of solving the Task Allocation problem based on a cost function. The tasks are represented with an AND/OR graph and the optimal task allocation is obtained solving the A* algorithm. The second layer, instead, handles the task execution and the respective failures.  However, if the system detects some errors, it is necessary to recalculate the optimal solution, which is a computationally demanding procedure. \\
	
These task assignment approaches assume that the time necessary for executing each task, either by the human or by the robot, is constant. This assumption is quite conservative and it is in sharp contrast with what happens in reality. In fact, it is quite unlikely that the human executes a given task in the same amount of time and this may lead to a strong inefficiency in the task allocation. Some works have addressed the variable task duration problem. In \cite{lou} the authors solve the variable task duration presenting a framework composed by two stages: the proactive scheduling stage and the reactive scheduling stage. In the first stage, the processing times are treated as Gaussian variables. Instead, the second stage is responsible of modifying the schedule to deal with uncertainties. In \cite{casalino} a scheduling architecture based on time Petri Nets is presented. The goal is to minimize the idle time of both human and the robot adapting the planned activities in a receding horizon. For this reason, the durations of the two agents are considered variable and fitted online with a Gaussian Mixture Model. However, this frameworks cannot handle the task assignment problem, which must be pre-solved. Moreover, the human and the robot are treated as two separated entities, without considering interactions and communication.	\\
	
In order to fully enable a human-robot collaboration, awareness and communication should be implemented both at the task execution level and at the task planning and scheduling level. Thus, it is important to create awareness about the real duration of the tasks executed by the human in order to enable the scheduler to replan the assigned tasks in order to maximize the efficiency of the collaboration. Furthermore, human and robot must be able to interact through the scheduler in order to make their collaboration smoother and more efficient. The human, because of its specific experience and expertise, can decide to execute a task that is assigned to the robot and the robot can decide to assign to the human one of its tasks (e.g. because it fails the execution). All these decisions should be communicated through the task scheduler.
	
In this paper we propose a novel framework for task assignment and scheduling for collaborative cells that is aware of the activity of the human and that allows the human and the robot to take decisions about the tasks they need to execute. The proposed framework is made up of two layers. Given a job to execute, an offline task assignment layer allocates the tasks to be executed by the human and the ones to be executed by the robot by building a nominal schedule. The scheduler layer, according to the real execution time of the human operator and to the decisions taken online by the human and by the robot, reschedules the tasks, possibly overriding the decisions taken by the task assignment layer, in order to improve the efficiency of the execution. 
	
The main contribution of this paper are:
		\begin{itemize}
			\item A novel adaptive framework for task assignment and scheduling that takes into account real execution time of the human and communication with human and robot for dynamic rescheduling 
			\item A strategy for dynamic rescheduling that is effective and computationally cheap, i.e. suitable for industrial applications, and that allows human and robot to communicate their needs to the scheduler. 
		\end{itemize}

 The paper is organized as follows: in Sec.~\ref{sec:problem} the task assignment and dynamic scheduling problem for a collaborative cell is detailed and in Sec.~\ref{section:task_assignment} an optimization procedure for solving the task assignment problem is proposed. In Sec.~\ref{sec:scheduler} an algorithm for dynamically scheduling the task of the robot and for allowing the human and the robot to communicate with the scheduler for changing the set of their assigned tasks is illustrated and in Sec.~\ref{sec:exps} an experimental validation of the proposed architecture is presented. Finally in Sec.~\ref{sec:conclusions} some conclusion and future work are addressed.
 
 



	\section{Problem Statement}\label{sec:problem}
	
        Consider a collaborative cell, where two agents, a human operator $H$ and  a robot $R$, have to execute a job $J$ together. The job can be split into a set of tasks\footnote{The choice of the specific technique for splitting a job into several tasks is out of the scope of this paper. Several strategies are available in the literature (see, e.g., \cite{johannsmeier} for assembly tasks.)} $(T_1,\dots ,T_N)$  and each task $T_i$ is associated with a \emph{nominal} execution time $t_i(A)$, where $A\in\{H,R\}$ represents the agent that executes the task. 

          The cell is endowed with a monitoring unit that, for a task $T_i$ assigned to the human,  estimates online the execution time. Several strategies for implementing the monitoring are available in the literature: sequential interval networks \cite{vo}, interaction  probabilistic movement primitives \cite{maeda}, Open-Ended Dynamic Time Warping (OE-DTW) \cite{maderna} to name a few.
		
	We assume that the tasks are independent, i.e. $T_i$ and $T_j$ can be executed in parallel for any $i,j =1, \dots, N$. Many interesting jobs are composed by independent tasks (e.g. serving parallel machines, simple assembly jobs).

We aim at designing a task assignment and dynamic scheduling architecture that:
		\begin{itemize}
			\item Builds optimal nominal task schedules for the human and for the robot, i.e. two task schedules such that, considering the nominal execution times, minimizes the job execution time and maximizes the parallelism between human and robot (i.e. minimizes idle time)
			\item Starting from the nominal task scheduling, reschedules the robot tasks according to the effective execution time detected by the monitoring unit and the decisions taken by the human and the robot for task swapping. The rescheduling aims at minimizing the execution time.
		\end{itemize}
		
		Continuously and automatically changing the order of the tasks assigned to the human can lead to confusion and poor efficiency of the operator \cite{stressindex}. Thus, we have chosen to reschedule online only the list of tasks assigned to the robot. The list of tasks assigned to the human changes only when necessary, namely when the human decides to execute a task assigned to the robot or when the robot cannot execute a task and asks for the help of the human. In this case, the lists of tasks assigned to the human is changed  minimally.
		
		The proposed task assignment and dynamic scheduling strategy can be represented by the architecture in Fig.~\ref{fig:framework}, where two main layers can be distinguished:
		\begin{figure}[t]
			\centerline{\includegraphics[width=\columnwidth]{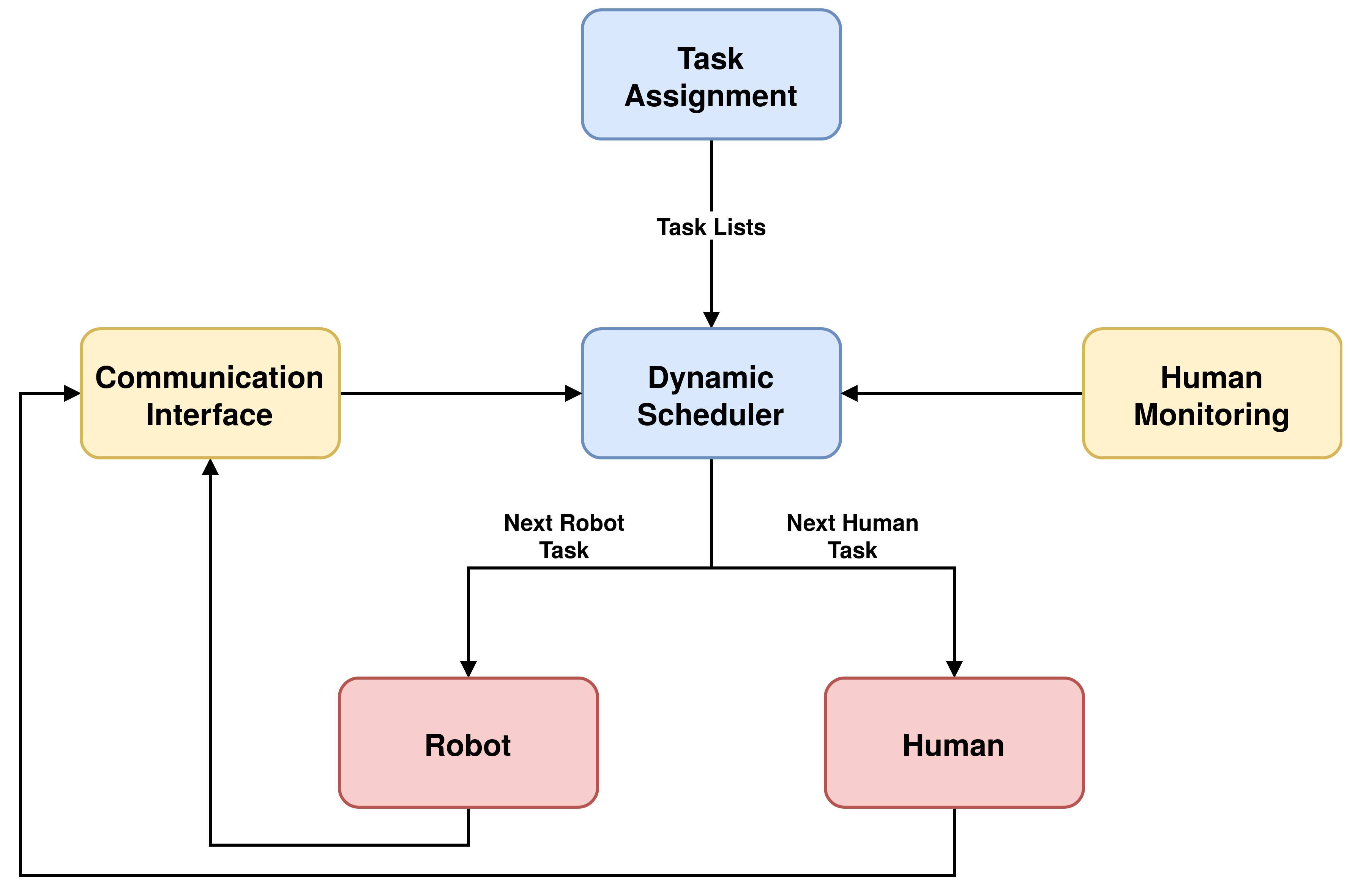}}
			\caption{The overall architecture. The blue blocks represent the two layers. The yellow blocks, instead, symbolize the strategies implemented to provide richer information to the Dynamic Scheduler. The red blocks represent the two agents.}
			\label{fig:framework}
		\end{figure}
		\begin{enumerate}
			\item \textbf{Task Assignment}. It is responsible of generating the initial nominal schedules for the robot and the human, based on the maximum parallelism criterion. 
			\item \textbf{Dynamic Scheduler}. It is responsible of scheduling the tasks, taking into account the real execution time and the requests coming from the human and from the robot.
		\end{enumerate}

	\section{Task Assignment Layer}
	\label{section:task_assignment}

The role of this layer is to build nominal task schedules for the human and for the robot starting from the nominal execution times of each task $T_i$. This is done by solving the following multi-objective Mixed Integer Linear Program:
	\begin{equation}
	\label{eq:MILP}
	\begin{array}[l]{ll}
	min_{x,c}\sum_{i=1}^N(w_{Ri} x_{Ri}+w_{Hi} x_{Hi})+c \\\\
	\text{subject to}\\\\
	x_{Ri}+x_{Hi}=1  \quad \forall i \in \{1,\dots , N\}\\\\
	\sum_{i=1}^Nt_{i}(a) x_{ai}\leq c \quad \forall a \in A\\\\
	\end{array}
	\end{equation}
	
        The terms $w_{Ri}, w_{Hi} > 0 $ represent the weights for executing task $T_i$ on behalf of the robot and of the human, respectively. The boolean variables $x_{Ri}, x_{Hi} \in \{0,1\}$ are detecting whether $T_i$ is assigned or not to the robot or to the human, respectively; $x=(x_{R1},\dots, x_{RN}, x_{H1},\dots, x_{HN})$ is the vector containing all the decision variables. Finally, $t_i(a)>0$ represents the nominal execution time of $T_i$ on behalf of agent $a\in A$ and $c>0$ denotes the cycle time.

The weight terms $w_{Ri}$ and  $w_{Hi}$ are exploited for encoding the skills of the collaborative agents in the execution of each task. The better the agent is at executing a task, the lower the corresponding weight. Very high weights are exploited for communicating to the task assignment algorithm that an agent is unsuitable for the execution of a task. In these terms job quality information can be considered. The first constraint guarantees that each task is assigned either to the robot or to the human. The second constraint maximizes the parallelism between the human and the robot. In fact, since all the terms in the quantity to minimize are positive, the optimization problem would tend to choose $c$ as small as possible and the lower bound for $c$ is given by the last constraint and corresponds to the maximum parallelization of the activities of the human and of the robot.
	
The outcome of the optimization problem \eqref{eq:MILP}  are $S_H$ and $S_R$, the set of tasks that have to be executed by the human and by the robot, respectively. The tasks in $S_H$ and $S_R$ are then organized in the increasing order of their indexes. This generates the nominal schedules, i.e. two ordered lists $L_H$ and $L_R$ containing the tasks that have to be sequentially executed by each agent.

	\section{Dynamic Scheduler}\label{sec:scheduler}

Starting from the nominal schedules $L_H$ and $L_R$, the goal of the dynamic scheduler is to take explicitly into account the variability of the human robot collaboration. When two humans are collaborating, if one gets slower the other tries to speed up its work in order to keep good performance of the team minimizing waiting times. Furthermore, difficulties are handled by communication. The most expert member of the team can decide to take some of the work or reorganize the work based on its experience. On the other side, when in trouble, the less experienced member of the team  asks the expert member for some help. The dynamic scheduler aims at reproducing this kind of behavior in human-robot collaboration in order to create an effective and intuitive cooperation.
	
This is achieved in two ways. First, by monitoring in real time the work of the human operator in order to estimate the real execution time and, if necessary, to reschedule the activities of the robot in order to avoid useless waiting times. Second, by enabling the human and the robot to communicate and take decisions about their activities through the scheduler. In particular, the robot allocates a task that it cannot execute to the human. The human can decide to execute the task that the robot is doing (because, e.g., from its experience, it feels that the robot is not executing the task in a proper way). Furthermore, the human can decide to re-allocate some its tasks to the robot. 
	
The dynamic scheduler is implemented according to the pseudo-code reported in Alg.~\ref{alg:dynamicscheduler}.
	
	\begin{algorithm}
		\caption{DynamicScheduler()}
		\label{alg:dynamicscheduler}
		\begin{algorithmic}[1]
			\State \textbf{Require:} $L_H$,$L_R$\label{algl:dsrequire}
			\State $End_R ,End_H \leftarrow false$ \label{algl:dsinitializeends}
			\State $T_R \leftarrow L_R(1)$; $T_H \leftarrow L_H(1)$ \label{algl:dsinitializetasks}
			\While{$(T_R\ne\emptyset$ \textbf{and} $T_H\ne\emptyset)$} \label{algl:dswhile}
			\State $End_R\leftarrow\mathbf{monitorR}(T_R)$ \label{algl:endr}
			\State $(L_R,End_H) \leftarrow\mathbf{reschedule}(T_H,L_R)$\label{algl:dsreschedule}
			\State $M_H\leftarrow \mathbf{read}_H()$, $M_R\leftarrow \mathbf{read}_R()$\label{algl:readmess}
			\State \resizebox{.85\hsize}{!}{$(End_H,End_R,L_H,L_R)= \mathbf{communication}(M_H, M_R, L_H,L_R)$} \label{algl:dscommunication}
			\If{$End_H$}
			$T_H=\mathbf{next} (T_H,L_H)$\label{algl:dsnextH}
			\EndIf
			\If{$End_R$}
			$T_R=\mathbf{next} (T_R,L_R)$\label{algl:dsnextR}
			\EndIf
			\EndWhile  
		\end{algorithmic}
	\end{algorithm}

The dynamic scheduler needs as  an input the nominal tasks schedules $L_H$ and $L_R$ (Line~\ref{algl:dsrequire}). It immediately sets to false the two variables $End_R$ and $End_H$ that identify when the task currently associated to the robot and to the human have been concluded (Line~\ref{algl:dsinitializeends}) and it assigns the first tasks of $L_R$ and of $L_H$ to the human and to the robot (Line~\ref{algl:dsinitializetasks}). The algorithm starts to loop until no more tasks are available for the robot and for the human (Line~\ref{algl:dswhile}). In the loop, the scheduler first checks if the robot finished the assigned task (Line~\ref{algl:endr}). The function $\mathbf{monitoR}(T_R)$ is application dependent and it verifies (e.g. by a camera, by a timeout procedure) if the task $T_R$ has been successfully accomplished. If this is the case, the $\mathbf{monitorR}$ function returns a $true$ value, $false$ otherwise. Furthermore, if the robot cannot succeed to execute $T_R$ (e.g. the task cannot be achieved after a predefined amount of time), the function $\mathbf{monitorR}$ generates a $delegate$ message $M_R$ and sends it to the scheduler. The $\mathbf{monitorR}()$  function can be implemented using standard procedures, available for robotic applications (see e.g. \cite{failure}). The human execution time is monitored and the list of tasks assigned to the robot is adapted in order to maximize in real time the parallelism. (Line~\ref{algl:dsreschedule}). Then the messages generated by the human and the robot are considered for task swapping (Line~\ref{algl:dscommunication}). Finally, the algorithm checks if the task assigned to the human and to the robot are over and, if it is the case, it assigns them the next task in the list (Lines~\ref{algl:dsnextH},~\ref{algl:dsnextR}). The function $\mathbf{next}(T,L)$ returns the task after $T$ in the list $L$ and, if $T$ is the last task of the list, it returns $\emptyset$. When both $T_R$ and $T_H$ are empty, then the job is over.
		
The rescheduling algorithm is represented in Alg.~\ref{alg:reschedule}.
	
	\begin{algorithm}
		\caption{Reschedule()}
		\label{alg:reschedule}
		\begin{algorithmic}[1]
			\State \textbf{Require:} $T_H$,$T_R$, $L_R$ \label{algl:rsrequire}
			\State $t_{res}\leftarrow \mathbf{monitorH}(T_H)$\label{algl:rsmonitor}
			\If{$t_{res}>t_R(R)$} \label{algl:rscheckresidualtime}
			\State $(pL_R,fL_R) \leftarrow \mathbf{split}(T_R,L_R)$\label{algl:rssplit}
			\State $fL_R^r \leftarrow\mathbf{fill}(fL_R,t_{res}-t_R(R))$\label{algl:rsknap}
			\State $L_R \leftarrow\mathbf{concat}(pL_R. L_R^r,fL_R/fL_R^r)$\label{algl:rsconcat}
			\EndIf
			\State $End_H \leftarrow \mathbf{not} t_{res}$\label{algl:rsendH}
			\State $\mathbf{return} (L_R,End_H)$\label{algl:rsreturn}
		\end{algorithmic}
	\end{algorithm}	
	
The algorithm requires as  input the tasks $T_H$ and $T_R$ that are currently  assigned to the human and to the robot and the current schedule $L_R$ of the tasks for the robot (Line~\ref{algl:rsrequire}). The activity of the human is monitored and the remaining time $t_{res}$ for the accomplishment of $T_H$ is estimated (Line~\ref{algl:rsmonitor}). The procedure \textbf{monitorH} can be implemented using several strategies available in the literature as, e.g., \cite{vo,maeda,maderna}. If $t_{res}$ is greater than the time necessary for the execution of task $T_R$, then some more tasks in $L_R$ may be executed  in parallel with $T_H$ and, therefore, the rescheduling procedure starts (Line~\ref{algl:rscheckresidualtime}). First, the list $L_R$ is split into two sub-lists: $pL_R$ contains $T_R$ and the previous tasks  while $fL_R$ contains the other tasks to be executed Line~\ref{algl:rssplit}. Then, from $fL_R$, a sublist $L_R^r$ of the tasks that can be executed in the extra time available $t_{res}-t_R(R)$ is generated Line~\ref{algl:rsknap}. The list $L_R^r$ can be generated using any version of the well-known knapsack algorithm \cite{knapsack} or more simple, job dependent, search techniques. Finally a new schedule for the robot tasks is built by  concatenating $pL_R$, with the list of the rescheduled tasks $fL_R^r$ and with the list of the remaining tasks to be executed Line~\ref{algl:rsconcat}. After the rescheduling, the $End_H$ is set to $true$ if $T_H$ is over or to $false$ otherwise (Line~\ref{algl:rsendH}) and, finally, the new schedule $L_R$ and the  $End_H$ variable are returned. 
		
\begin{figure}[t]
			\centerline{\includegraphics[width=\columnwidth]{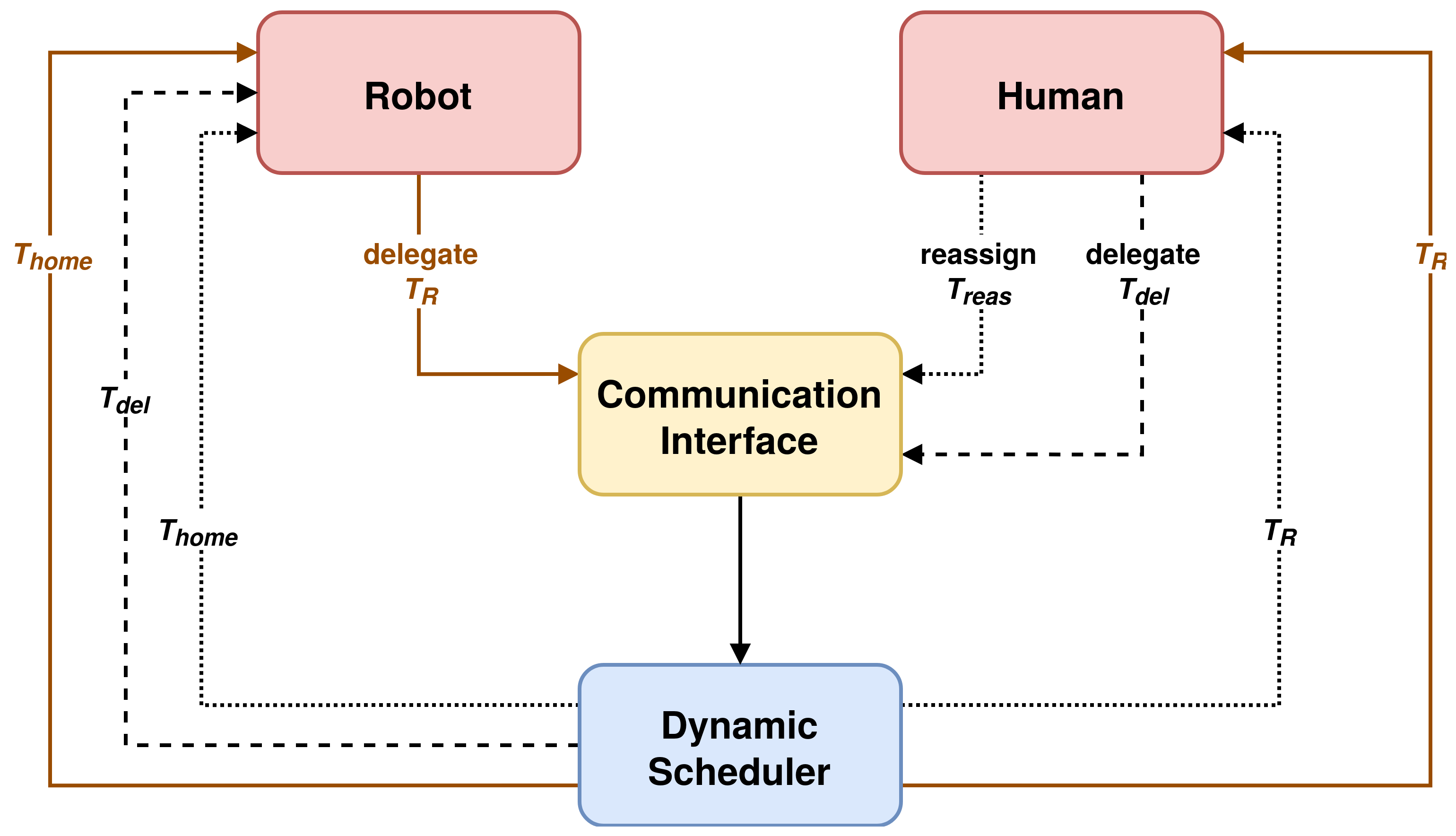}}
			\caption{Communication Layout. The dotted lines indicate the ``reassign" message coming from the human and the consequent task scheduling. The dashed lines indicate the ``delegate" message coming from the human with the following scheduling to the robot. Finally, the brown lines indicate the ``delegate" message communicated by the robot.}
			\label{fig:messages}
		\end{figure}

		During the execution of the job the human and the robot can generate messages in order to communicate to the Dynamic Scheduler the intention or need to swap their tasks. A detailed representation of this communication layout is shown in Fig.~\ref{fig:messages}.
	In particular, the message $M_R$ sent by the robot can be either empty or containing the value ``delegate $T_{R}$''  and it is generated by the \textbf{monitorR} function when the robot cannot succeed in executing the assigned task. The message $M_H$ can be either empty or it can assume two values: ``reassign $T_{reas}$'' or ``delegate $T_{del}$''. The first message is generated when the human decides to execute the task that the robot is executing (because, e.g., the robot is not doing the assigned work properly or in the best way). The second message is generated when the human decides to delegate some of the tasks in $L_H$ to the robot. This message has an argument, that specifies the task to be delegated. The human can enter the messages through a proper, job dependent, input interface. The messages generated by the human and by the robot are handled by Alg.~\ref{alg:communication}.
	
	\begin{algorithm}
		\caption{Communication()}
		\label{alg:communication}
		\begin{algorithmic}[1]
			\State \textbf{Require:} $M_H$, $M_R$, $T_R$, $L_H$,$L_R$ \label{algl:corequire}
			\If{$M_H=reassign(T_{reas})$ } 
			\State $End_R\leftarrow true$\label{algl:coendr}
			\State $End_H\leftarrow true$\label{algl:coendh}
			\If{$T_{reas} = T_R$} \label{algl:cochecktreas}
				\State$L_R\leftarrow\mathbf{push}(T_{home},L_R)$\label{algl:copushhome1}
			\EndIf
			\State
			$L_R\leftarrow\mathbf{delete}(T_{reas},L_R)$\label{algl:codelete3}
			\State  $L_H\leftarrow\mathbf{push}(T_{reas},L_H)$\label{algl:copushRH}
			\ElsIf{$M_H=delegate(T_{del})$ \textbf{and} $\mathbf{exRobot}(T_{del})$}\label{algl:cocheckexrob}
			\State $L_H\leftarrow\mathbf{delete}(T_{del},L_H)$\label{algl:codelete1}
			\State   $L_R\leftarrow\mathbf{push}(T_{del},L_R)$\label{algl:copushdelR}
			\EndIf
			\If{$M_R=delegate(T_{R})$ \textbf{and} $\mathbf{exHuman}(T_R)$}\label{algl:cocheckexhum}
			\State $End_R\leftarrow true$\label{algl:coendr2}
			\State $L_R\leftarrow\mathbf{delete}(T_R,L_R)$\label{algl:codelete2}
			\State $L_R\leftarrow\mathbf{push}(T_{home},L_R)$\label{algl:copushhomeR}
			\State   $L_R\leftarrow\mathbf{push}(T_R,L_H)$\label{algl:copushRH2}
			\EndIf
			\State $\mathbf{return}$ $End_H$, $End_R$, $L_H$, $L_R$\label{algl:coreturn}
		\end{algorithmic}
	\end{algorithm}	
	
The algorithm requires the messages generated by human and robot $M_R$ and $M_H$ and the current schedules $L_R$ and $L_H$ and the task $T_R$ currently assigned to the robot (Line~\ref{algl:corequire}). The message $M_H$ is the first to be handled in order to give priority to the decisions taken by the operator. If the human decides to execute a task that was initially assigned to robot (i.e. $M_H = ``reassign$ $T_{reas}"$), both the end of the task variables are set to true. In this way, in Alg.~\ref{alg:dynamicscheduler}, the robot and the human will be assigned a new task (Lines~\ref{algl:coendr} and \ref{algl:coendh}). If the reassigned task is the one that the robot is executing (Line~\ref{algl:cochecktreas}), an homing task $T_{home}$ is put as the next task in the robot schedule (Line~\ref{algl:copushhome1}). The task $T_{reas}$ is then deleted from $L_R$ (Line~\ref{algl:codelete3}) while the task $T_R$ is pushed in the first position of the human schedule (Line~\ref{algl:copushRH}). In this way, the Alg.~\ref{alg:dynamicscheduler} will allocate $T_{home}$, that will send the robot to a safe home position, and $T_R$, the task the human has decided to execute, as the next tasks for the robot and the human. If the human decides to allocate a task $T_{del}\in L_H$ (i.e. $M_H =``delegate$ $T_{del}"$) and the task is executable by the robot (Line~\ref{algl:cocheckexrob}), then $T_{del}$ is deleted from $L_H$ (Line~\ref{algl:codelete1}) and transferred into the task schedule of the robot (Line~\ref{algl:copushdelR}). 
		If the robot detects that it cannot fulfill the assigned task (i.e. $M_R=``delegate"$) and if the task it is trying to accomplish is executable by the human operator (Line~\ref{algl:cocheckexhum}), then the end of the task variable for the robot is set to $false$ to force Alg.~\ref{alg:dynamicscheduler} to allocate the next task to the robot. $T_R$ is deleted from the robot schedule $L_R$ (Line~\ref{algl:codelete2}) and inserted in the schedule of the human  (Line~\ref{algl:copushRH2}). Furthermore, an homing mission $T_{home}$ is added as the next task for the robot.
		Finally the procedure returns the updated end of task variables and schedules (Line~\ref{algl:coreturn}).
		The procedure \textbf{exRobot} and \textbf{exHuman} exploit prior information about the job and the tasks (e.g. the weights $w_{Ri}$ and $w_{Hi}$  in \eqref{eq:MILP}) to detect if a task can be executed by the robot or by the human.

	\section{Experiments}\label{sec:exps}
	The proposed two-layers framework has been experimentally validated in a collaborative assembly job consisting of storing $4$ plastic shapes, fixing $3$ little PCBs and positioning a big PCB and a wooden bar. The human operator cooperated with a Kuka LWR 4+, a 7-DoF collaborative robot, at whose end-effector a 3D printed tool was attached. This tool allowed picking the objects using magnets.
	To monitor the human task execution we used a Kinect V2 RGB-D Camera with the official APIs for the skeleton tracking, while to evaluate the remaining task time we implemented the OE-DTW algorithm (see \cite{tormene}) to the operator wrists positions. This algorithm compares an incomplete input time series with a reference one, returning as output the fraction of the reference series that corresponds to the input. This fraction corresponds to the percentage of completion of the task $\%_{compl}$ and it is used to estimate the remaining time as following:
	\begin{equation}
	    t_{res} = (1-\%_{compl})t_{i}(H)
	\end{equation}
	
\begin{figure}[!tbp]
		\centering
		\begin{minipage}[b]{0.49\columnwidth}
			\includegraphics[width=\columnwidth]{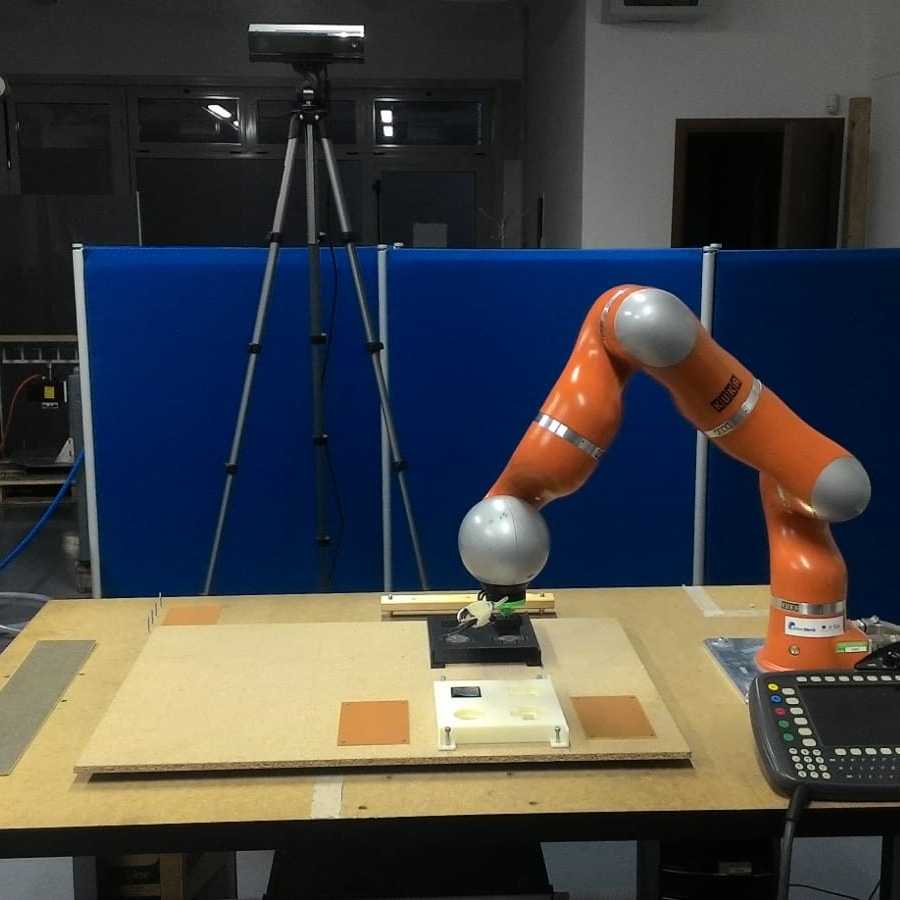}
		\end{minipage}
		\hfill
		\begin{minipage}[b]{0.49\columnwidth}
			\includegraphics[width=\columnwidth]{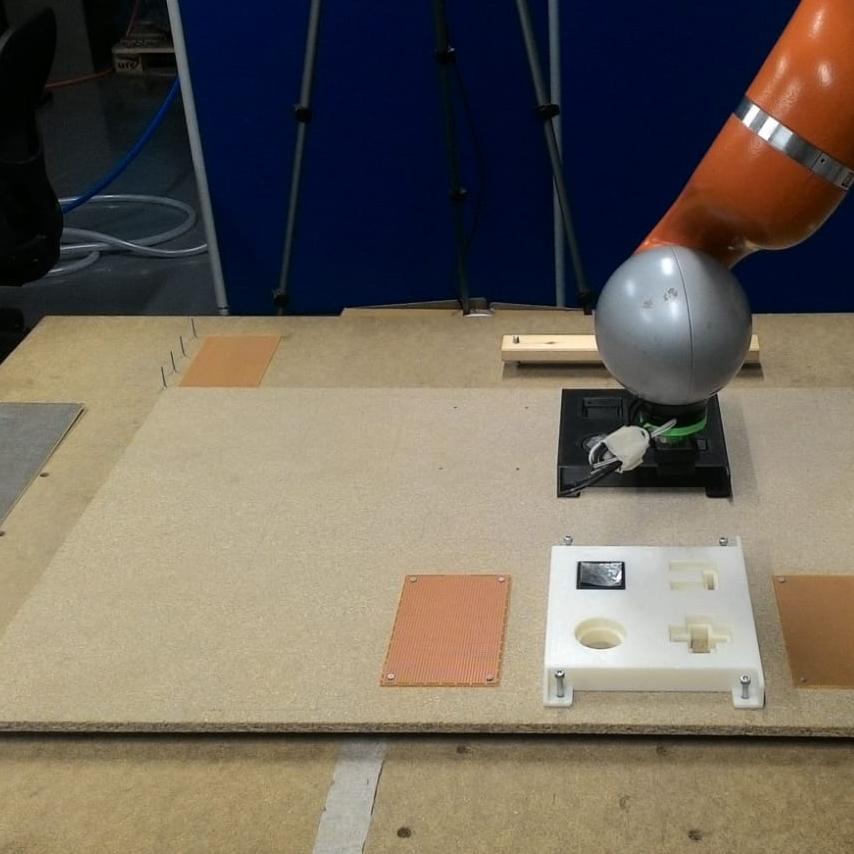}
		\end{minipage}
		\caption{Setup of the experiment. The two images show all the equipment used during the experiments. Moreover it is possible to note the presence of two specular shape selectors. The black one is fixed and it represents the area where the shapes were picked. The white selector, instead, is the one placed and screwed during the assembly job. It represents the area where the shapes were placed.}
		\label{fig:photo_setup}
	\end{figure}

	For the communication interface, instead, we developed an HMI that allowed the operator to generate the wanted messages with the keyboard of the computer. A complete setup of the experiment is shown in Fig.~\ref{fig:photo_setup}.
	
	All the software components were developed using ROS Melodic Morenia meta-operating system and they ran on a Intel(R) Core(TM) i7-4700MQ CPU @ 2.40GHz with Ubuntu 18.04.
	The optimization problem was implemented using JuMP\cite{jump},
	a modelling language embedded in Julia programming language, and solved with Cbc solver\cite{cbc}.
		
	We divided the collaborative assembly job in $11$ different tasks, $(T_1, ..., T_{11})$. Each task consists of one or more actions, this is due to create a more natural collaboration and communication. In fact, sometimes it could be more efficient to reassign and delegate a sequence of actions instead of a single one. A detailed description of the actions to be performed for each task is provided in Tab.~\ref{table:task_description}. The collaboration of the two agents started after the tasks $T_1$ and $T_2$ were completed.

\begin{table}[!htbp]
	\centering
	\caption{Tasks Action}
	\label{table:task_description}
	\newcolumntype{Y}{>{\centering\arraybackslash}X}
	\begin{tabularx}{\linewidth}{YX}
		\toprule
		\textbf{Task Index} & \textbf{Description} \\
		\midrule
		1 & Pick\&Place of a shape sorter. \newline Pick\&Place of the screws. \\
		\midrule
		2 & Screwing of the shape sorter.\\
		\midrule
		3 & Pick\&Place of the first PCB. \newline Pick\&Place of the screws. \newline
		Screwing. \\
		\midrule
		4 & Pick\&Place of the second PCB. \newline Pick\&Place of the screws. \newline
		Screwing. \\
		\midrule
		5 & Pick\&Place of the third PCB. \newline Pick\&Place of the screws. \newline
		Screwing. \\
		\midrule
		6 & Pick\&Place of the bigger PCB. \\
		\midrule
		7 & Pick\&Place of a cross shape. \\
		\midrule
		8 & Pick\&Place of a circular shape. \\
		\midrule
		9 & Pick\&Place of a U shape. \\
		\midrule
		10 & Pick\&Place of a square shape. \\
		\midrule
		11 & Pick\&Place of a wooden bar. \\
		\bottomrule
	\end{tabularx}
\end{table}

In the experiments we considered as input for the Task Assignment Layer (see Sec.~\ref{section:task_assignment}) the data in Tab.~\ref{table:optimization_data}. The weights $w_{Ri}$ and $w_{Hi}$ have been estimated taking into account the distance in the shared workspace between the task components and the agent and the intrinsic capability of each agent to accomplish the task (e.g. placing the screws is more difficult for the robot, because it requires a high precision).
It is important to notice that all the tasks durations are less than equal to one. This is due to the fact that all the durations were normalized respect to the maximum nominal one, which is $t_{robot, 6} = 40\,sec$. The nominal durations were precalculated by measuring many times the time required for each agent to perform the tasks and taking the average value.

\begin{table}[!htbp]
	\centering
	\caption{Task Assignment Data}
	\label{table:optimization_data}
	\newcolumntype{Y}{>{\centering\arraybackslash}X}
	\begin{tabularx}{\linewidth}{YYYYY}
		\toprule
		\textbf{Task Index} & $\bm{w_{Ri}}$ & $\bm{t_{i}(R)}$ & $\bm{w_{Hi}}$ & $\bm{t_{i}(H)}$\\
		\midrule
		1 & 0.6 & 0.500 & 0.14 & 0.375\\
		\midrule
		2 & 0.4 & 0.375 & 0.06 & 0.250 \\
		\midrule
		3 & 0.8 & 0.875 & 0.2 & 0.625 \\
		\midrule
		4 & 0.8 & 0.875 & 0.2 & 0.625 \\
		\midrule
		5 & 0.8 & 0.875 & 0.2 & 0.625 \\
		\midrule
		6 & 0.9 & 1.000 & 0.1 & 0.375 \\
		\midrule
		7 & 0.3 & 0.350 & 0.5 & 0.250 \\
		\midrule
		8 & 0.3 & 0.350 & 0.5 & 0.250 \\
		\midrule
		9 & 0.3 & 0.350 & 0.5 & 0.250 \\
		\midrule
		10 & 0.3 & 0.350 & 0.5 & 0.250 \\
		\midrule
		11 & 0.2 & 0.250 & 0.9 & 0.750 \\
		\bottomrule
	\end{tabularx}
\end{table}

The solution of the optimization problem was a first nominal schedule that minimize the cost function, which was composed by the following two lists:
\begin{itemize}
	\item $L_H = (1,2,3,4,5, 6)$
	\item $L_R = (7,8,9,10, 11)$
\end{itemize}

	Starting from the output of the Task Assignment Layer, the Dynamic Scheduler was then initialized and tested in the collaborative assembly scenario. A complete video of the experiments is attached\footnote{\url{https://youtu.be/48pH6MpSytM}}. The first part of the video is dedicated to the first experiment where the two agents, the human and the robot, execute exactly the expected tasks, namely the \textit{''nominal schedule"}. Initially the robot is in idle, because the tasks $T_1$ and $T_2$ are preparatory for the collaborative job, which is not started yet. After the human confirms the completion of that task, the Dynamic Scheduler allows the robot to start executing the parallel tasks. During the execution of $T_3$ the monitoring strategy is activated and the framework starts to estimate the remaining time ($T_{res}$). After the robot executes $T_7$ and $T_8$, the estimated $T_{res}$ becomes lower than the required time of all the other robot tasks (see Alg.~\ref{alg:reschedule}, Line~\ref{algl:rscheckresidualtime}) and no other tasks can be executed by the robot, e.g. $fL^r_R = \emptyset$ (see Alg.~\ref{alg:reschedule}, Line~\ref{algl:rsknap}). At this point, the human slows down while performing the screwing action, causing an increment in the estimated remaining time. Thanks to the adopted rescheduling strategy the vector of the rescheduled tasks is filled with the other available parallel tasks, $fL^r_R = \{T_8,T_9\}$, and the robot starts executing the first one.
	
	In the second experiment, the communication strategy is exploited. The operator starts to execute the nominal schedule concluding $T_1$. However after placing the screws, he sends a message ``delegate $T_2$" to the Dynamic Scheduler. The start of the collaborative job is anticipated and the robot executes the screwing action while the operator is free to proceed his schedule. Please note that during the experiments it was not possible to execute e real screwing, for this reason we simulate the task by placing the robot over the screws and applying a rotation. After a while, thanks to his the great expertise, the operator realizes that the robot will place in a wrong way the \textit{``U''} shape, which is the task $T_9$. For this reason, the operator executes $T_9$ instead of the robot and communicate the message ``reassign $T_9$" to the Dynamic Scheduler. At this point the Dynamic Scheduler deletes that task from the schedule of the robot and the robot executes the next one, $T_{10}$.

	\section{Conclusion and Future Works} \label{sec:conclusions}
	In this paper we propose a two-layers framework for task assignment and scheduling for collaborative cells. Exploiting the tracking of the human body, the framework takes into account real execution time of the human to adopt an effective dynamic rescheduling strategy. Moreover, in order to take advantage of the expertise of the operator and deal with possible errors, the presented framework allows the real-time communication between the human and the robot. The experimental evaluation shows the effectiveness of the framework and its applicability in a real industrial scenario, with parallel tasks involved.
	
        Future work aims at removing the assumption of independency of the tasks, that can be quite restrictive for some application. Furthermore, in order to
further improve the cooperation between the human operator and the robot, we will investigate how the rescheduling and the communication system affect the human cognitive workload, taking more explicitly into account all the dimensions of  job quality. This will lead to the possibility of optimize in real-time the industrial process, not only minimizing the idle times and maximizing the parallelism but also improving the well-being of the human operators.

\bibliographystyle{splncs03_unsrt}

\bibliography{bib}

\end{document}